\documentclass[runningheads]{llncs}

\usepackage[T1]{fontenc}
\usepackage{graphicx}
\usepackage{booktabs}
\usepackage{multirow}
\usepackage{colortbl}
\usepackage[pagebackref,breaklinks,colorlinks]{hyperref}

\usepackage{wrapfig}
\usepackage{orcidlink}
\usepackage{xr-hyper}
\usepackage{algorithm}
\usepackage{algpseudocode}
\usepackage{amsmath,bm}
\usepackage{bbm}
\usepackage{amsfonts}
\usepackage{caption}
\usepackage{subfig}

\usepackage{graphicx}
\usepackage{color} 

\definecolor{Gray}{gray}{0.9}
\definecolor{Better}{rgb}{0.18, 0.407, 0.266}
\definecolor{Worse}{rgb}{0.35, 0.35, 0.35}
\definecolor{drakgreen}{rgb}{0.38, 0.67, 0.38}
\definecolor{drakpurple}{rgb}{0.38, 0.27, 0.61}
\definecolor{granate}{rgb}{0.64, 0.16, 0.16}

\usepackage[euler]{textgreek}
\usepackage{arydshln}
\usepackage{marvosym}

\newcommand{\ww}{\mathbf{w}}
\newcommand{\KK}{\mathbf{K}}

\newcommand{\mm}{\mathbf{m}}

\newcommand{\grad}{\mathbf{g}}

\newcommand{\pp}{\mathbf{p}}                     
\newcommand{\W}{\mathbf{W}}                     
\newcommand{\yy}{\mathbf{y}} 
\newcommand{\real}{\mathbb{R}}                   
\newcommand{\vv}{\mathbf{v}}                     
\newcommand{\ttt}{\mathbf{t}}                    
\newcommand{\temp}{\tau}                         
\newcommand{\domain}[1]{$\mathcal{D}_\text{#1}$} 

\begin{document}

\title{Few-Shot, Now for Real: Medical VLMs Adaptation without Balanced Sets or Validation}
\titlerunning{Medical VLMs Adaptation without Balanced Sets or Validation}

\author{
Julio Silva-Rodríguez\inst{1}\textsuperscript{,\Letter}
\and
Fereshteh Shakeri\inst{1,2}
\and
Houda Bahig\inst{2} 
\and \\
Jose Dolz\inst{1,2}
\and
Ismail {Ben Ayed}\inst{1,2}
}

\authorrunning{J.~Silva-Rodríguez et al.}

\institute{
\inst{1}ÉTS Montréal \\ 
\Letter {\tt \small \email{julio-jose.silva-rodriguez@etsmtl.ca}} \\
\inst{2}Centre de Recherche du Centre Hospitalier de l’Université de Montréal (CRCHUM) 
}

\index{Silva-Rodríguez, Julio}
\index{Shakeri, Fereshteh}
\index{Bahig, Houda}
\index{Dolz, Jose}
\index{Ben Ayed, Ismail}

\maketitle          

\begin{abstract}

Vision-language models (VLMs) are gaining attention in medical image analysis. These are pre-trained on large, heterogeneous data sources, yielding rich and transferable representations. Notably, the combination of modality-specialized VLMs with few-shot adaptation has provided fruitful results, enabling the efficient deployment of high-performing solutions. However, previous works on this topic make strong assumptions about the distribution of adaptation data, which are unrealistic in the medical domain. First, prior art assumes access to a balanced support set, a condition that breaks the natural imbalance in disease prevalence found in real-world scenarios. Second, these works typically assume the presence of an additional validation set to fix critical hyper-parameters, which is highly data-inefficient. This work challenges these favorable deployment scenarios and introduces a realistic, imbalanced, validation-free adaptation setting. Our extensive benchmark across various modalities and downstream tasks demonstrates that current methods systematically compromise their performance when operating under realistic conditions, occasionally even performing worse than zero-shot inference. Also, we introduce a training-free linear probe that adaptively blends visual and textual supervision. Detailed studies demonstrate that the proposed solver is a strong, efficient baseline, enabling robust adaptation in challenging scenarios. Code is available: \url{https://github.com/jusiro/SS-Text} .

\end{abstract}

\section{Introduction}
\label{sec:intro}

The recent advancements in deep learning have yielded remarkable outcomes to enhance computer-aided medical image analysis \cite{mediasurv}. Nevertheless, these have been classically hindered by the necessity of using large, labeled datasets for training highly specialized solutions \cite{Chen2022}. Currently, foundation models are driving a paradigm shift, enabling more data-efficient, robust solutions \cite{Moor2023,foundMed}. Particularly, specialized contrastive vision-language models (VLMs) are being developed in the primary medical image modalities, i.e., radiology \cite{convirt,MedCLIP}, histology \cite{PLIP,CONCH} or retina \cite{FLAIR}. These models are pre-trained on large datasets through text supervision, which enables the leveraging of heterogeneous data sources that encompass different tasks and domains. Thanks to such pre-training, they exhibit remarkable adaptability to various downstream tasks. For example, VLMs enable zero-shot predictions by leveraging text knowledge into the learned joint multi-modal space and have shown astonishing robustness to domain drifts \cite{radford2021learning}. However, recent studies have underscored that such zero-shot performance highly depends on the target concept's frequency \cite{udandarao2024zeroshot}, and domain \cite{forgotten_domain_generalization} during pre-training. This motivates the integration of supervisory signals for the target tasks, ideally in a data-efficient manner, i.e., using small numbers of labeled samples, also known as few-shot adaptation. 

Few-shot adaptation is of special interest in medical applications due to the scarcity of data resources or when tackling low-prevalence diseases. From a practical standpoint, a practitioner would gather a few examples for each category, e.g., 4, 8, or 16, and efficiently adapt a pre-trained model. Indeed, VLMs are strong few-shot learners \cite{radford2021learning}, particularly in the efficient black-box scenario \cite{ouali2023black}, i.e., transferring the embedded representations. A recent body of literature has focused on improving this transfer using the so-called Adapters \cite{gao2021clip,zhang2021tip,yu2023task,lin2023crossmodal,clap24,lp24}, which combine visual and textual supervision. Consequently, due to its success in natural image domains, recent work in \cite{shakeri2024few} has deployed this setting in modern modality-specialized medical VLMs. However, we argue that the scenarios explored in natural domains are hardly directly applicable to medical image analysis, where data access presents more challenging constraints.

\begin{figure*}[htb]
    \begin{center}
        \begin{tabular}{lll}

         \includegraphics[width=.49\linewidth]{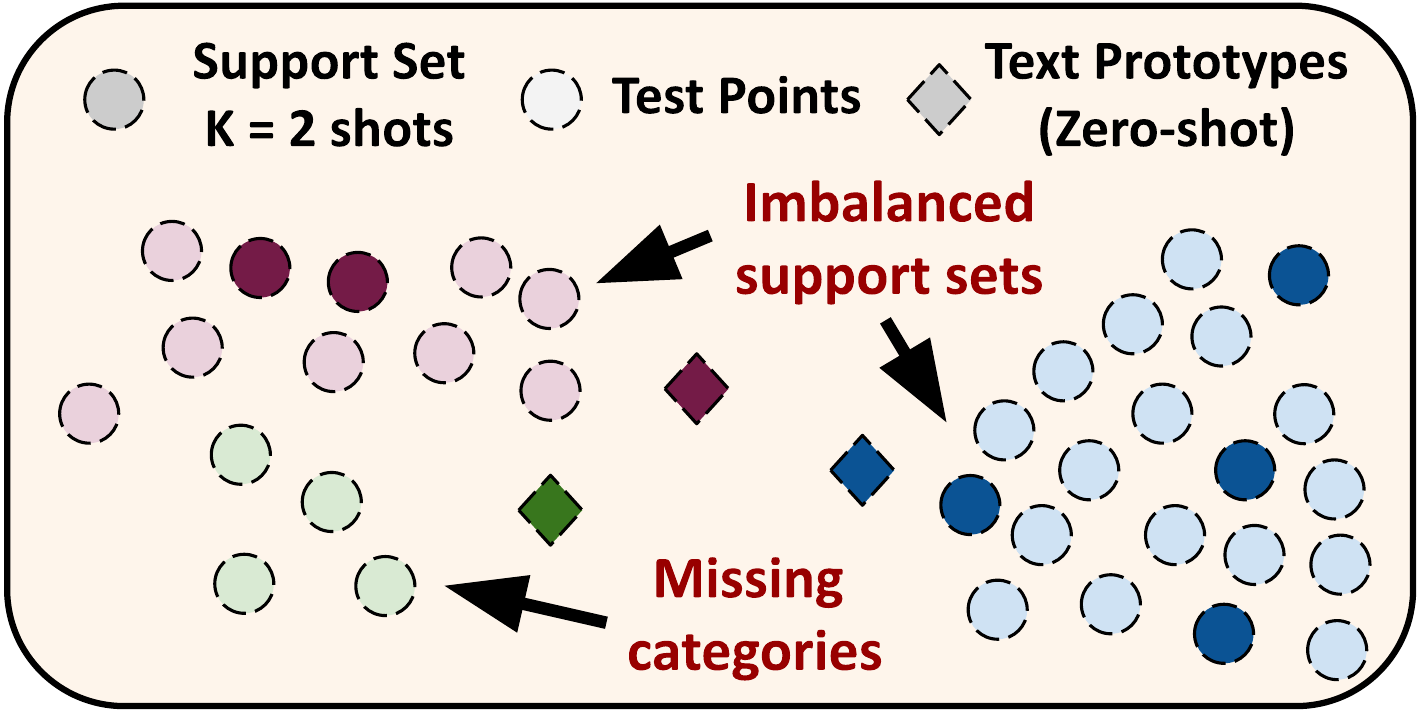} & 
         & \includegraphics[width=.49\linewidth]{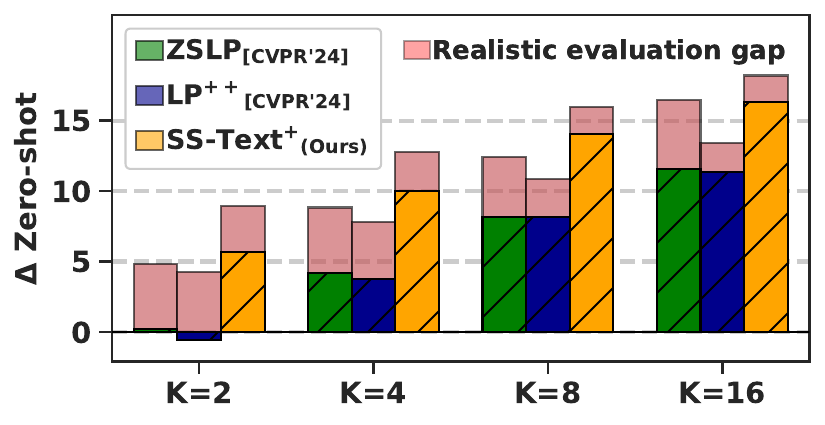} \\

          \multicolumn{1}{c}{\textbf{(a) Realistic few-shot sets}} & & \multicolumn{1}{c}{\textbf{(b) Performance drop}} \\

        \end{tabular}
        \caption{\textbf{Realistic adaptation of medical VLMs}: disease prevalence might produce imbalanced support sets or missing categories during low-shot adaptation (a), producing severe performance detriments in SoTA black-box Adapters (b).}
        \label{fig:motivation}
    \end{center}
\end{figure*}

First, these works assume access to a balanced support set for adaptation, such that the number of shots is homogeneous across categories. However, this strategy is unrealistic in medical domains, where disease prevalence is naturally imbalanced. For example, in tumor grading tasks \cite{silva2020going} or diabetic retinopathy grading \cite{messidor}, gathering examples for the severe disease stages is typically challenging. Indeed, in low-data regimes, there may be potentially missing categories in the support set (the least prevalent ones). Thus, evaluating such few-shot Adapters using balanced support sets provides an overoptimistic, biased assessment of the potential of these techniques in real-world scenarios. Note that this is not an anecdotal problem, but its effects are being observed in recent literature. For example, the authors of Quilt \cite{ikezogwo2023quiltm} reported better performance in low-data but balanced regimes than using the whole dataset for adaptation when transferring histology VLMs. Moreover, if evaluated in realistic scenarios (see Fig.~\ref{fig:motivation}(b)), current SoTA methods suffer significant performance degradation.

Second, current few-shot adaptation strategies suffer from data-inefficient model selection strategies, i.e., how critical hyper-parameters and learning schedules are fixed. Early works even leveraged significant data sources for this task \cite{gao2021clip,zhang2021tip}, a limitation smoothed by ulterior literature \cite{lin2023crossmodal,zhang2021tip}. Still, current medical benchmarks assume a few-shot validation set \cite{shakeri2024few}, which doubles the required samples for adaptation. Again, such a scenario poses an additional burden in medical applications. Therefore, we argue that few-shot Adapters should rely uniquely on a single support set and incorporate appropriate model selection strategies. Indeed, as we later demonstrate in Fig. \ref{fig:efficiency_val}(b), using such additional validation data for training brings larger performance gains. \\ 

\noindent The main contributions of this paper can be summarized as:
\begin{itemize}
    \item[$\circ$] We highlight the limitations of current few-shot Adapters for medical VLMs, which often degrade its performance (Fig.~\ref{fig:motivation}(b)) in realistic scenarios that account for real-world disease prevalence and constraints to data access.
    \item[$\circ$] Hence, we introduce two validation-free scenarios for assessing its few-shot adaptation with imbalanced support sets and potential missing categories in the low-data regime (Fig.~\ref{fig:motivation}(a)).
    \item[$\circ$] We propose SS-Text$^{+}$, a training-free, constraint linear probe that adaptively leverages imbalanced visual and textual information for each category, providing more robust performance than the current state-of-the-art (SoTA). 
\end{itemize}

\section{Related Work}
\label{sec:rw}

\noindent\textbf{\textit{Transfer learning in VLMs}.} Initial efforts were devoted to Prompt Learning \cite{zhou2022coop,kgcoop23}, which learns a set of input tokens to optimize the best textual prototypes. However, current literature has underscored its inefficiency compared to black-box Adapters \cite{gao2021clip,zhang2021tip,lp24,clap24}, which directly operate over pre-computed features. While their performance is comparable, they are orders of magnitude faster \cite{lp24,shakeri2024few}. We follow up on \cite{shakeri2024few} and focus on assessing these efficient Adapters.

\noindent\textbf{\textit{Black-box Adapters}.} Current methods are linear probes that combine textual and visual information. Indeed, a simple linear probe with proper textual-based initialization is already a strong baseline adaptation \cite{clap24}. More complex methods can be categorized into ensemble-based \cite{gao2021clip,zhang2021tip,lp24,shakeri2024few}, which combine visual and textual similarities into the final output, and constraint-based \cite{lin2023crossmodal,clap24}, which enforce the learned probes to remain close to the initial textual class embeddings. For example, the CrossModal linear probe \cite{lin2023crossmodal} employs textual embeddings as additional support samples for logistic regression, while CLAP \cite{clap24} applies explicit penalties. Our proposed solver belongs to the second family of methods and provides a training-free, flexible solver designed for imbalanced scenarios.

\section{Methods}
\label{sec:background}

\subsection{Contrastive vision-language models}
\label{ssec:zero-shot}

\noindent\textbf{\textit{Zero-shot}.} Contrastive VLMs project data points into an $\ell_{2}$-normalized $D$-dimensional shared embedding space, yielding the corresponding visual, ${\vv} \in \real^{D \times 1}$, and text, ${\ttt} \in \real^{D \times 1}$, embeddings. Given a pre-computed image feature and class-wise prototypes, $\W=(\ww_c)_{1 \leq c \leq C}$, with $\ww_c \in \real^{D \times 1}$, and $C$ the number of target categories, probabilities can be computed as:
\begin{align}
\label{eq:probs}
    \phantom{,}p_c(\W)
    = \frac
    {\exp( \vv^\top \ww_{c} / \temp)}
    {\sum_{j=1}^{C} \exp( \vv^\top \ww_j / \temp)},
\end{align}
where $\temp$ is a pre-trained temperature parameter, $\vv^\top \ww$ is the cosine similarity, and $\pp(\W)=(p_c(\W))_{1 \leq c \leq C}$ corresponds to the predicted probabilities vector. Contrastive VLMs allow zero-shot predictions, i.e., no need to learn $\W$, by embedding a textual description for each label. Concretely, given a set of $J$ textual embeddings for each target category, $\{\{\ttt_{cj}\}_{j=1}^{J}\}_{c=1}^{C}$, the zero-shot prototypes are the average of the text embeddings for each class, $\ttt_{c}=\frac{1}{J}\sum_{j=1}^{J}\ttt_{cj}$. These class weights are then applied to Eq.~\eqref{eq:probs}, i.e., $\W^*=(\ttt_c)_{1 \leq c \leq C}$.

\subsection{Towards realistic few-shot adaptation scenarios}
\label{ssec:realistic_setting}

Let us define an adaptation set, \domain{adapt}$=\{(\vv_n,\yy_n)\}_{n=1}^{N}$, composed of labeled embedded visual examples. Note that $\yy$ is the one-hot encoding of the label space, $\mathcal{Y}=\{1, 2, ..., C\}$. Typically, in a few-shot support set, $N$ takes small values, defined by the number of examples available for each category, e.g., $K \in \{1, 2, 4, 8, 16\}$, such that the total number of samples is $N=K\times C$. In the following, we detail the standard adaptation scenario explored in prior art, and two more challenging scenarios we propose for addressing real-world constraints:
\begin{itemize}
    \item[$\circ$] \textbf{\textit{i}) Standard}. The number of shots for each category, $\KK=(K_c)_{1 \leq c \leq C}$, is constant, such that $K_c=N/C$. However, this assumption does not account for the realistic imbalance that naturally arises from variable disease prevalence in specific demographics or institutions.
    \item[$\circ$] \textbf{\textit{ii}) Realistic}. Let us define a label-marginal distribution among categories, $\mm=(m_c)_{1 \leq c \leq C} \ , \ \text{s.t.} \sum\mm=1$. For a specific $K$, the support set comprises $N=K\times C$ samples, each from a category sampled from the label-marginal distribution, i.e., $y\sim\mm$. In this scenario, the number of samples from a given category might be variable and even zero if $m_c$ and $K$ are small, which is a challenging scenario with potentially missing categories.
    \item[$\circ$] \textbf{\textit{iii}) Relaxed}. This intermediate scenario assumes that at least one labeled example can be retrieved for each category of interest, i.e. $K_c\geq1 \ \forall c \in C$.

\end{itemize}

\subsection{SS-Text: training-free, text-informed linear probe}
\label{ssec:sstext}

Multi-modal black-box Adapters involve linear probes that combine visual and textual information \cite{lin2023crossmodal,clap24}. These can be modeled from a constrained optimization perspective, where the weights matrix, $\W$ in Eq.~\eqref{eq:probs}, is adjusted minimizing cross-entropy loss, but encouraged to remain near the textual class prototypes:
\begin{equation}
\label{eq:objective}
    \phantom{.}\min_{\W} \
    \mathcal{L}(\W) = - \frac{1}{N} \sum\limits_{i=1}^{N} \sum\limits_{c=1}^{C}
    y_{ic} \ \text{ln}(p_{ic}(\W)) + \frac{\lambda}{2} \sum\limits_{c=1}^{C} ||\ww_c - \ttt_c||_{2}^{2}.
\end{equation}
This objective is typically optimized via gradient descent \cite{clap24}. However, due to the scarcity of adaptation data, its global minimum might be suboptimal when generalizing to test data. Current literature controls such overfitting by learning procedures based on gradient-based optimizers, which necessitate a fine-grained hyper-parameter search, e.g., for iterations, and learning rate schedule, thereby negating the validation-free scenario explored in this work. To address such limitation, we leverage our previously introduced training-free solver, SS-Text, which was recently presented in \cite{fca25} for its speed in adaptation. The objective in Eq.~\eqref{eq:objective} is evaluated as the sum of two terms, $\mathcal{L} = g_1 + g_2$, such that:
\begin{align}
   & g_1 = - \frac{1}{N} \sum\limits_{i=1}^{N} \sum\limits_{c=1}^{C}
    y_{ic} \ (\vv^\top \ww_{c}  / \temp) + \frac{\lambda}{2} \sum\limits_{c=1}^{C} ||\ww_c - \ttt_c||_{2}^{2} , \label{g1} \\
   & g_2 = \frac{1}{N} \sum\limits_{i=1}^{N} \text{ln} \left( \sum\limits_{j=1}^{C}
    \text{exp}(\vv^\top \ww_{j}  / \temp) \right). \label{g2}
\end{align}
The solution of Eq.~\eqref{eq:objective} can be approximated by solely addressing the first term, $g_1$, which is related to a hard-label assignment of the support embeddings. Independent of the value of $\lambda > 0$, $g_1$ is convex w.r.t. each $\ww_c$, as it is the sum of linear and convex functions. Particularly, its minimum is:
\begin{equation}
\label{eq:solver}
\ww_c^{g_1} = \phantom{.}\arg\min_{\ww_c} \ \frac{\partial g_1}{\partial \ww_c} = \frac{1}{\lambda N \temp} \sum\limits_{i=1}^{N} y_{ic} \vv + \ttt_c.
\end{equation}
This solution is a linear combination of scaled visual and textual embeddings for each class. Its closed-form makes it convenient for validation-free deployments. However, it may not yet handle the imbalanced, realistic scenarios addressed in this work, as visual and textual information are weighted equally across all categories. To overcome such a limitation, we propose \textbf{SS-Text$^{+}$}, a modified version of the solution in \cite{fca25} for this specific challenge.

\textbf{First}, the general multiplier lambda is fixed adaptively class-wise, such that $\lambda_c = 1/(K_c\temp)$. Thus, the effect of text knowledge decreases as more support data becomes available for a specific category. Note that this choice accommodates the solution to the realistic, imbalanced scenario with missing categories. For example, if no supervision is provided for one particular category ($K_c=0$), the obtained prototype will still rely on the available textual information.

\textbf{Second}, to avoid prototypes overlapping due to insufficient supervision, e.g., when missing categories and relying primarily on text supervision, we propose a post-processing stage that refines the overall solution. Concretely, the resultant prototypes are distanced one each other such that $\ww_c^{*} = \ww_c^{g_1} - \frac{\grad_c}{||\grad_c||},$ where $\grad_c=\ww_c^{g_1} - \frac{1}{C-1} \sum_{c'\in \mathcal{Y}_{c}'} \ww_{c'}^{g_1}$, such that $\mathcal{Y}_c'=\{c' \in \mathcal{Y} \ | \ c'\neq c\}$. This term can be interpreted as repelling $\ww_c^{g_1}$ from the prototypes of the other categories.

\section{Experiments}
\label{sec:experiments}

\subsection{Setup}

\noindent\textbf{\textit{Medical vision-language models}.} Three modalities are tackled, as in \cite{shakeri2024few}. \textbf{Histology}: CONCH \cite{CONCH} is used, which deploys a customized larger-scale ViT-B/16 visual backbone. \textbf{Ophtalmology:} FLAIR \cite{FLAIR}, the expert-knowledge guided VLM for retina imaging, is selected. \textbf{Chest X-ray (CXR)}: CONVIRT \cite{convirt} pre-trained on MIMIC \cite{mimic} is used. FLAIR and CONVIRT are composed of fine-tuned BioClinicalBERT text encoders \cite{bioclinicalbert}, and the vision encoder is ResNet-50.

\noindent\textbf{\textit{Downstream tasks}} include tissue/disease classification or grading in balanced (\textit{b}) and imbalanced (\textit{u}) scenarios. \textbf{Histology}: three different organs are included: colon in NCT-CRC \cite{kather2018100} (\textit{u}), prostate Gleason grading in SICAPv2 \cite{silva2020going} (\textit{u}), and SkinCancer \cite{kriegsmann2022deep} (\textit{u}). \textbf{Ophtalmology}: diabetic retinopathy using MESSIDOR \cite{messidor} (\textit{u}), myopic maculopathy staging in MMAC \cite{mmac} (\textit{u}), and disease detection in FIVES \cite{fives} (\textit{b}), are included. \textbf{CXR}: five CheXpert \cite{irvin2019chexpert} (\textit{b}) categories, as in \cite{MedCLIP}, 19 categories in NIH-LT \cite{nih,nih_lt} (\textit{u}), and pneumonia types \cite{covid1,covid2} (\textit{b}) are assessed.

\noindent\textbf{\textit{Adaptation methods}.} Relevant gradient-based black-box Adapters are leveraged as baselines. These are trained in a validation-free setting, i.e., using fixed learning hyper-parameters: training for 300 steps using full-batch SGD optimizer with a momentum of $0.9$ and an initial learning rate of $0.1$ with a cosine-scheduler decay. For vanilla linear probing, we follow ZS-LP initialization \cite{clap24}. Also, the recent text-informed SoTA CLAP \cite{clap24} and LP++ \cite{lp24,shakeri2024few} are integrated. For LP++, the learning rate is data-driven, as proposed by the authors. Finally, previous relevant multi-modal Adapters are assessed: CLIP-Adapter \cite{gao2021clip} ($\alpha_{\text{clipad}}=0.2$), TIP-Adapter \cite{zhang2021tip}($\alpha_{\text{tip}}$, $\beta_{\text{tip}}=1.0$) training-free and fine-tuned (f), TaskRes \cite{yu2023task}($\alpha_{\text{taskres}}=1.0$), and CrossModal linear probe \cite{lin2023crossmodal}.

\noindent\textbf{\textit{Evaluation protocol}.} Support sets with $K \in \{1, 2, 4, 8, 16\}$ are sampled from train data as described in Section \ref{ssec:realistic_setting}. The label-marginal distribution, $\mm$, is estimated from the whole train subset by measuring label frequencies. Class-wise balanced accuracy (ACA) is employed as a figure of merit, as recommended in \cite{metrics}. All results are the average across 20 different random seeds for sampling.

\subsection{Results}

\begin{figure*}[t!]
    \begin{center}
        \begin{tabular}{lll}

         \includegraphics[width=.4\linewidth]{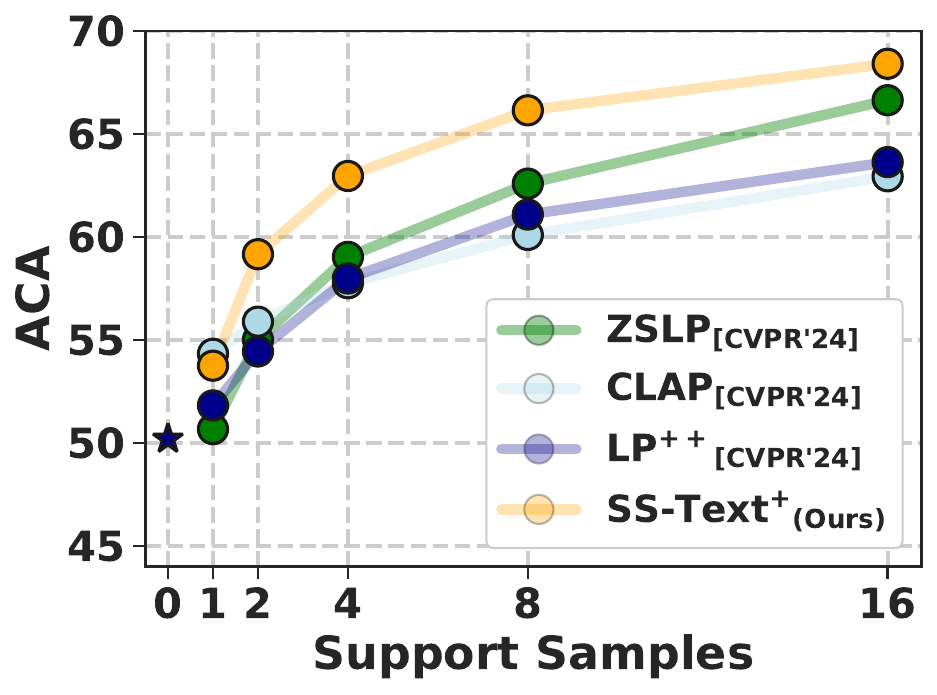} & 
         & \includegraphics[width=.4\linewidth]{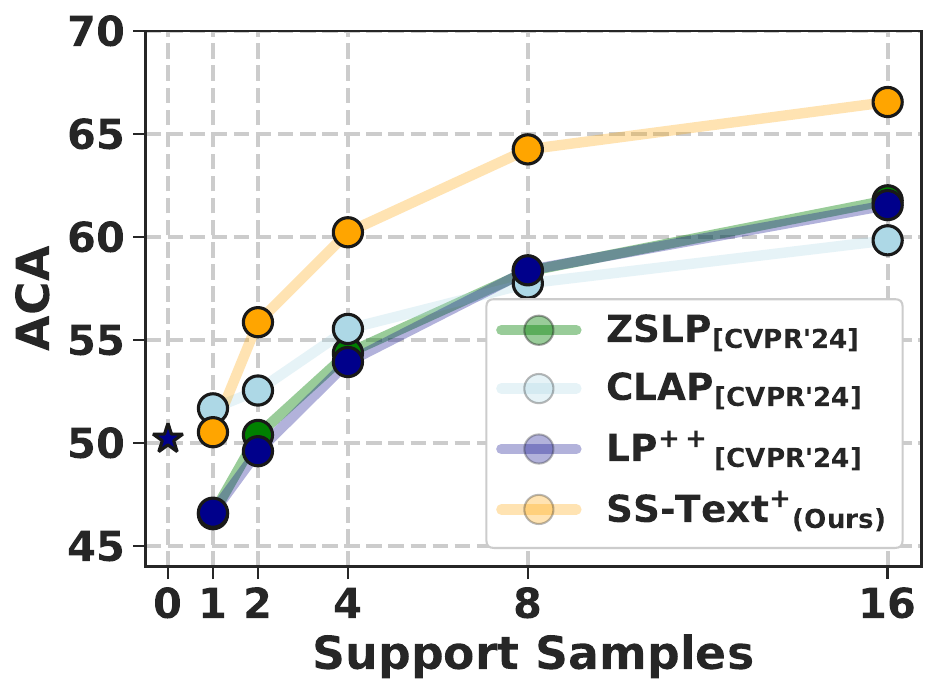} \\

          \multicolumn{1}{c}{\textbf{(a) Standard setting}} & & \multicolumn{1}{c}{\textbf{(b) Realistic scenario}} \\

        \end{tabular}
        \caption{\textbf{Results across shots}: comparison of SoTA black-box Adapters.}
        \label{fig:fewshot}
    \end{center}
\end{figure*}

\noindent\textbf{\textit{Results across shots}.} Results for the most recent SoTA are depicted in Fig. \ref{fig:fewshot} for the standard (a) and realistic (b) scenarios. First, it is worth noting that the proposed SS-Text$^{+}$ outperforms relevant SoTA in the validation-free scenario. Second, all methods consistently drop the performance when considering realistic support sets. Note that some methods degrade below Zero-shot (no adaptation) in the low-shot scenario. In contrast, our method is more robust to this decay.

\noindent\textbf{\textit{Detailed results per modality}.} Table \ref{detailed} introduces the performance for the three explored scenarios per modality. Results are reported with K=4, a low-shot regime showcasing the effects of the missing classes. For example, vanilla linear probing (ZSLP) degrades nearly $-3.7$ in the relaxed setting and $-4.6$ in the realistic scenario. This performance drop is alleviated in text-informed solutions, such as LP++, which suffers a drop of $-1.9$ and $-4.1$, respectively. Note that the gap is larger in the realistic scenario, where some methods perform below zero-shot for certain modalities, e.g., LP++ in ophthalmology. In contrast, SS-Text$^{+}$ is more robust and the best solution across all scenarios.

\begin{table}[htb]
\setlength{\tabcolsep}{2.5pt}
\scriptsize
\centering
\caption{\textbf{Results per modality and type of support set} with $K=4$ shots.}
\label{detailed}
    \begin{tabular}{lcccccccccccc}
    \toprule
    & \multicolumn{4}{c}{\textbf{(a) Standard}} & \multicolumn{4}{c}{\textbf{(b) Relaxed}} & \multicolumn{4}{c}{\textbf{(c) Realistic}} \\ \cmidrule(lr){2-5}\cmidrule(lr){6-9}\cmidrule(lr){10-13}
    \multicolumn{1}{c}{\multirow{1}{*}{Method}} & Hist.     & Opht.    & CXR    & \textbf{Avg.} & Hist.     & Opht.    & CXR    & \textbf{Avg.} & Hist.     & Opht.    & CXR    & \textbf{Avg.}  \\ 
    \midrule
    Zero-shot \cite{radford2021learning}               & 61.7 & 56.8 & 31.8 & 50.1 & 61.7 & 56.8 & 31.8 & 50.1 & 61.7 & 56.8 & 31.8 & 50.1   \\
    CLIP-Adapt \cite{gao2021clip}                      & 78.1 & 60.1 & 42.3 & 60.2 & \underline{74.0} & 58.2 & 39.5 & \underline{57.2} & \underline{73.0} & 56.7 & 38.8 & \underline{56.2}   \\
    TIP-Adapt \cite{zhang2021tip}                      & 77.4 & 61.7 & 43.5 & 60.8 & 38.0 & 37.4 & 36.8 & 37.4 & 38.0 & 38.2 & 36.8 & 37.7   \\
    TIP-Adapt(f) \cite{zhang2021tip}                   & \textbf{82.0} & \textbf{63.6} & \underline{43.7} & \textbf{63.1} & 65.6 & 52.6 & 37.7 & 52.0 & 65.6 & 46.8 & 37.6 & 50.0   \\
    TaskRes \cite{yu2023task}                          & 74.7 & 60.5 & 41.9 & 59.0 & 71.4 & 56.7 & 38.0 & 55.3 & 71.0 & 54.4 & 37.7 & 54.4   \\
    CrossModal \cite{lin2023crossmodal}                & 71.0 & 62.2 & 44.0 & 59.1 & 70.9 & 60.0 & 40.1 & 57.0 & 70.8 & 56.3 & 39.4 & 55.5   \\
    ZSLP \cite{clap24}                                 & 74.7 & 60.5 & 41.9 & 59.0 & 71.4 & 56.7 & 38.0 & 55.3 & 71.0 & 54.4 & 37.7 & 54.4   \\
    CLAP \cite{clap24}                                 & 66.9 & 62.6 & 43.7 & 57.8 & 67.5 & \underline{61.0} & 40.0 & 56.2 & 67.4 & \underline{59.7} & \underline{39.5} & 55.5   \\
    LP++ \cite{lp24}                                   & 70.2 & 61.0 & 42.7 & 58.0 & 68.8 & 59.3 & \underline{40.2} & 56.1 & 68.6 & 54.9 & 38.3 & 53.9   \\
    \rowcolor{Gray}SS-Text$^{+}$ (\textit{Ours})       & \underline{81.2} & \underline{63.2} & \textbf{44.4} & \underline{63.0} & \textbf{76.3} & \textbf{62.8} & \textbf{44.2} & \textbf{61.1} & \textbf{75.0} & \textbf{61.7} & \textbf{44.0} & \textbf{60.2}   \\ 
    \bottomrule
    \end{tabular}
\end{table}

\begin{figure*}[htb]
    \begin{center}
        \begin{tabular}{lll}

         \includegraphics[width=.42\linewidth]{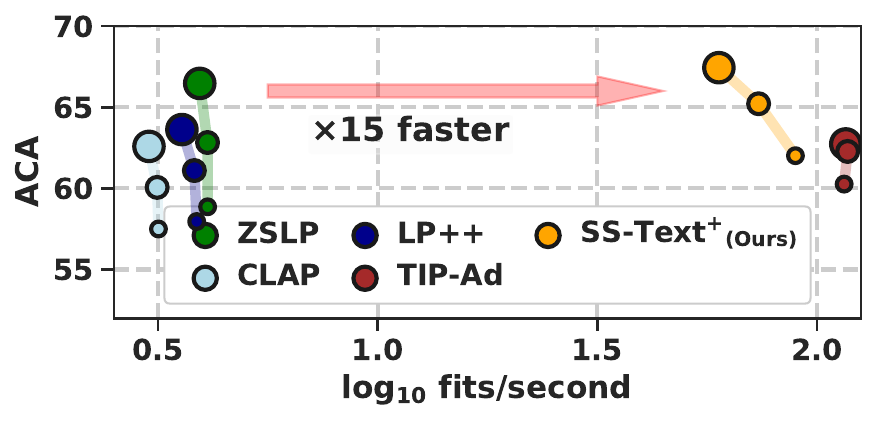} & 
         & \includegraphics[width=.42\linewidth]{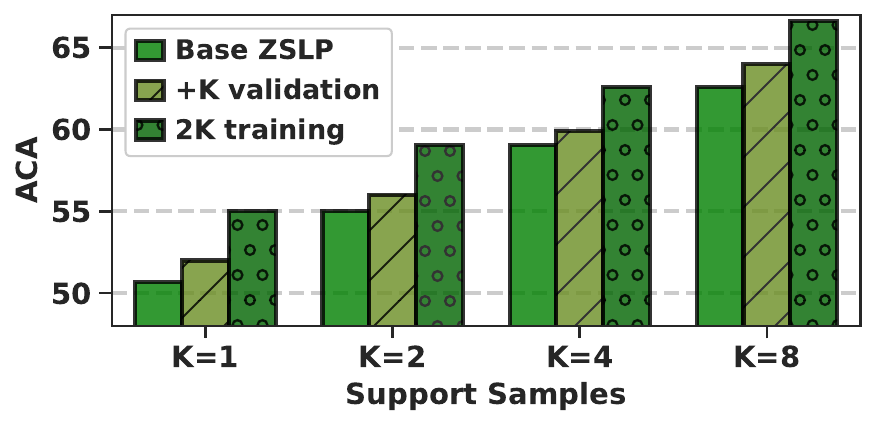} \\

          \multicolumn{1}{c}{\textbf{(a) Efficiency}} & & \multicolumn{1}{c}{\textbf{(b) How to use the validation set?}} \\

        \end{tabular}
        \caption{\textbf{Efficiency and model selection studies}. Results in the standard scenario. In (a), The dot size indicates the number of shots, i.e., $K \in \{4, 8, 16\}$.}
        \label{fig:efficiency_val}
    \end{center}
\end{figure*}

\noindent\textbf{\textit{Efficiency analysis}.} Fig. \ref{fig:efficiency_val}(a) shows the efficiency of SS-Text$^{+}$. Note that gradient-based linear probes are already highly efficient due to their black-box nature, especially compared to prompt learning as assessed in \cite{shakeri2024few}. Notably, SS-Text$^{+}$ is $\geq15\times$ faster, which showcases its utility in high-latency applications.

\noindent\textbf{\textit{Is it worth using a few-shot validation set}?} Using the vanilla linear probe, Fig. \ref{fig:efficiency_val}(b) provides evidence that the performance gains from using few-shot validation sets are smaller than leveraging this data for training a validation-free approach that follows a fixed setting and does not require model selection.

\noindent\textbf{\textit{Hyper-parameter analysis}.} Table \ref{ablation_studies}(a) evaluates the proposed task-adaptive text regularization weight $\lambda_c$. Our configuration, which considers the specific amount of supervision received from each category, provides a robust benefit across shots compared to non-adaptive alternatives or performance-driven choices as proposed in CLAP \cite{clap24}. Also, Table \ref{ablation_studies}(b) shows the benefit of the repelling prototype post-processing, which provides a $+0.7$ gain in accuracy.

\noindent\textbf{\textit{Long-tail training}.} Popular imbalanced deep learning mechanisms are re-balancing, augmentation, or decoupled learning techniques (see \cite{nih_lt}). Due to the black-box nature of few-shot adaptation, the latter ones do not apply to our scenario, and the first, e.g., class weighting (cw), is not applicable if missing categories. Table \ref{ablation_studies}(c) reports the positive effect of re-balancing techniques over vanilla linear probing in the explored relaxed setting. Still, its performance falls behind the proposed text-informed, training-free solution (see Table \ref{ablation_studies}(b)).

\begin{table}[htb]
\setlength{\tabcolsep}{2pt}
\scriptsize
\centering
\caption{\textbf{Ablation studies} in SS-Text$^{+}$ and long-tail learning techniques.}
\label{ablation_studies}
\begin{tabular}{ cccc }
    \begin{tabular}{c}
        \begin{tabular}{lccccc}
            \toprule
            \multicolumn{1}{l}{\multirow{1}{*}{Method \ \ $K\rightarrow$}} & $1$     & $2$    & $4$    & $8$    & $16$   \\
            \midrule
            \multicolumn{6}{l}{Text regularization fixed for all tasks.} \vspace{0.5mm}  \\
            \hdashline\noalign{\vskip 0.5ex}
            $\lambda=0.1/\temp$          & 49.1 & 54.7 & 59.9 & 63.7 & 66.4   \\
            $\lambda=1.0/\temp$          & 48.2 & 53.7 & 57.6 & 60.6 & 61.5   \\
            $\lambda=10/\temp$           & 27.7 & 29.4 & 29.3 & 29.1 & 29.5   \\
            \midrule
            \multicolumn{6}{l}{Ours: task-adaptive regularization value.}  \vspace{0.5mm} \\
            \hdashline\noalign{\vskip 0.5ex}
            $\lambda_c$ as in \cite{clap24}          & 49.6 & 55.1 & 60.1 & 63.4 & 65.5   \\
            \rowcolor{Gray}$\lambda_c=1/(K_c\temp)$  & \textbf{50.5} & \textbf{55.9} & \textbf{60.2} & \textbf{64.2} & \textbf{66.5}   \\
            \bottomrule
        \end{tabular} \\
    \textbf{(a) $\lambda_c$ in SS-Text$^{+}$} (realistic setting)
    \end{tabular}
    & & &
    \begin{tabular}{c}
        \begin{tabular}{lcccc}
        \toprule
        \multicolumn{1}{c}{\multirow{1}{*}{Method}} & Hist.     & Fundus    & CXR    & \textbf{Avg.}   \\ 
        \midrule
        SS-Text$^{+}$ w/o repel                           & \textbf{83.2} & 67.0 & 47.2 & 65.8 \\
        \rowcolor{Gray}SS-Text$^{+}$                      & \textbf{83.2} & \textbf{67.5} & \textbf{48.9} & \textbf{66.5}   \\
        \bottomrule
        \end{tabular} \\ 
    \textbf{(b) SS-Text$^{+}$ Components} (K=16, realistic) \vspace{2mm} \\
        \begin{tabular}{lcccc}
        \toprule
        \multicolumn{1}{c}{\multirow{1}{*}{Method}} & Hist.     & Fundus    & CXR    & \textbf{Avg.}   \\ 
        \midrule
        ZSLP \cite{clap24}                         & 77.2 & 63.3 & 45.1 & 61.9   \\
        \hspace{1mm} w/ cw                         & 79.2 & 65.4 & 47.5 & 64.0   \\
        \hspace{1mm} w/ cw+LDAM\cite{ldam}         & \textbf{79.6} & \textbf{65.6} & \textbf{47.7} & \textbf{64.3}   \\
        \bottomrule
        \end{tabular} \\ 
    \textbf{(c) Long-tail strategies} (K=16, $K_c\geq1$) \\
    \end{tabular} \\
 
\end{tabular}
\end{table}

\section{Conclusions}
\label{sec:conclusion}

Several aspects have been introduced that should be considered when adapting medical vision-language models in the few-shot data regime. These include a more realistic design of the support set and validation-free model selection pipelines. In this context, the proposed training-free SS-Text$^{+}$ is a strong baseline for measuring progress in the field.

\begin{credits}
\subsubsection{\ackname} This work was funded by the Natural Sciences and Engineering Research Council of Canada (NSERC) and the Montréal University Hospital Research Center (CRCHUM). We also thank Calcul Quebec and Compute Canada.

\end{credits}

\bibliographystyle{splncs04}
\bibliography{refs}

\end{document}